# A Framework for Shape Analysis via Hilbert Space Embedding

Sadeep Jayasumana[1,2], Mathieu Salzmann[1,2], Hongdong Li[1], and Mehrtash Harandi[1,2]

[1]Australian National University, Canberra      [2]NICTA, Canberra[*]

sadeep.jayasumana@anu.edu.au

## Abstract

*We propose a framework for 2D shape analysis using positive definite kernels defined on Kendall's shape manifold. Different representations of 2D shapes are known to generate different nonlinear spaces. Due to the nonlinearity of these spaces, most existing shape classification algorithms resort to nearest neighbor methods and to learning distances on shape spaces. Here, we propose to map shapes on Kendall's shape manifold to a high dimensional Hilbert space where Euclidean geometry applies. To this end, we introduce a kernel on this manifold that permits such a mapping, and prove its positive definiteness. This kernel lets us extend kernel-based algorithms developed for Euclidean spaces, such as SVM, MKL and kernel PCA, to the shape manifold. We demonstrate the benefits of our approach over the state-of-the-art methods on shape classification, clustering and retrieval.*

## 1. Introduction

The ability of humans to utilize shape as a prominent cue to identify objects has resulted in the popularity of shape analysis in a wide variety of computer vision applications including object recognition [1], image segmentation [4], activity recognition [24], biomedical image analysis [5] and human-computer interaction [8].

In computer vision, the term *shape* refers to the geometric information of an object that is invariant to translation, scale and rotation [5]. Since the works of Kendall [10] and Bookstein [3], significant progress has been made in the area of shape analysis. While a number of methods have been proposed to encode shapes as mathematical objects [28], Kendall's shape manifold remains the most popular and widely used shape representation.

In Kendall's framework, a 2D shape represented by $m$ landmarks can be treated as a point in the complex projective space $\mathbb{C}P^{m-2}$ [10]. Although an appropriate metric on this space can be obtained in closed form via Procrustes analysis, the space itself, known as the *shape manifold*, is a Riemannian manifold with nontrivial geometry. Due to the nonlinear geometry of the shape manifold, popular shape recognition and retrieval methods are limited to either employing simple techniques such as nearest-neighbors [1, 13], or modeling the manifold-valued data with a complex Bingham distribution [5], which may not reflect the true sample distribution.

A common approach to cope with the nonlinearity of a manifold consists in approximating the manifold-valued data with its projection to a tangent space at a particular point on the manifold (e.g., the mean of the data). For the shape manifold, this projection can be achieved while simultaneously preserving rotation invariance [5]. However, while the resulting space is indeed Euclidean, such a tangent space approximation can significantly distort the original manifold-valued data, especially in regions far from the point around which the tangent space was computed.

In Euclidean spaces, the success of many computer vision algorithms arises from their use of kernel methods [21]. It would therefore seem natural and attractive to extend this approach to manifold-valued data and embed the manifold in a high dimensional Reproducing Kernel Hilbert Space (RKHS), where Euclidean geometry applies. Such a mapping however, requires a kernel function, which, according to Mercer's theorem, must be positive definite. One could think of replacing the Euclidean distance in the popular Gaussian radial basis function with the metric on the manifold to obtain a kernel function. However, a kernel function derived in this manner is *not* positive definite in general.

In this paper, we introduce the *Procrustes Gaussian kernel*, a provably positive definite kernel on the shape manifold. Being positive definite, this kernel function allows us to embed the shape manifold in a high dimensional Hilbert space. To the best of our knowledge, this represents the first embedding of the shape manifold in a Hilbert space, thus letting us generalize the use of kernel methods to the shape manifold. The advantage of such an embedding is twofold: First, it enables the use of well established recognition methods that require linear geometry on the nonlinear shape manifold. Second, as evidenced by kernel methods on $\mathbb{R}^n$, embedding a lower dimensional space in a higher dimensional one helps identifying complex patterns in a given data distribution.

More specifically, we make use of the full Procrustes dis-

---

[*]NICTA is funded by the Australian Government as represented by the Department of Broadband, Communications and the Digital Economy and the ARC through the ICT Centre of Excellence program.

This work is funded in part through two ARC Discovery grants: DP120103896 and DP130104567.

tance [10, 5] as the distance measure on the shape manifold, and show that it gives rise to a positive definite Gaussian kernel. We exploit this kernel in four different algorithms; support vector machines (SVM), multiple kernel learning (MKL), kernel principal component analysis and kernel $k$-means. To demonstrate the benefits of the Hilbert space embedding of shapes obtained with our kernel, we tackle the tasks of shape classification, clustering and retrieval. Our experimental evaluation shows that our manifold-based kernel methods outperform tangent space approaches and other shape analysis baselines on these tasks.

## 2. Related Work

Shape analysis has been widely studied in computer vision [28]. Over the years, different shape representations have been proposed, such as landmarks [10], level sets [4], distance transformed contours [26] and others [28]. Most of these representations yield nonlinear shape spaces where Euclidean geometry does not apply. In this paper, we study the popular and successful shape manifold introduced by Kendall, where a shape is represented by a finite number of landmarks.

The lack of Euclidean structure of the shape manifold has made it impossible to utilize well known algorithms, such as SVM, to perform shape recognition while accounting for the manifold structure. Therefore, existing approaches have instead focused on nearest neighbor classification [13, 1] and on designing better distance measures for such classifiers [1]. Alternatively, recognition has been performed by attempting to model the probability distribution of shape data on the complex unit sphere with a complex Bingham distribution [5, 8], or rotation invariant distributions [8]. Unfortunately, such distributions are often too restrictive to accurately model the true data distribution. Furthermore, our experience with Euclidean spaces strongly suggests that better classifiers can be obtained by exploiting kernel methods.

The term *kernel* is used somewhat loosely in the literature of shape analysis without special attention to positive definiteness. The use of a non-positive definite Chamfer kernel was proposed in [26, 29] in the context of shape detection. In [8], a rotation invariant kernel was introduced and shown to be effective for shape recognition. The notion of invariant kernel functions was also studied in a more general context in [7] and [25]. As acknowledged in most of the above-mentioned works, none of these kernels are known to be positive definite. A number of positive definite kernels, derived from learnt distances, were proposed in [6]. Our work differs from theirs since we use the standard full Procrustes distance. Kernels have been also used for density estimation and modeling implicit surfaces in [4] and [15], respectively. These works, however, do not use kernels to embed shapes in a Hilbert space. Recently, we introduced positive definite kernels on a different manifold [9].

According to Mercer's theorem, only positive definite kernels define valid RKHSs. Positive definiteness of kernels is also required for convexity of many popular learning algorithms such as SVM [18] and MKL [23]. While attempts have been made to exploit non-positive definite kernels [27, 17], many of them have the drawback of altering the eigenvalues of the kernel matrix to artificially make it positive definite [27].

In this paper, we introduce a provably positive definite kernel for the shape manifold. This allows us to exploit existing powerful kernel methods for shape analysis.

## 3. The Shape Manifold

In Kendall's formalism, a 2D shape is initially represented as an $m$-dimensional complex vector, where $m$ is the number of landmarks that denote the shape. The real and imaginary parts of each element of the vector encode the $x$ and $y$ coordinates of a landmark, respectively. Translation and scale invariances are achieved by subtracting the mean from the vector and scaling it to have unit norm. The vector $\mathbf{z}$ obtained in this manner is dubbed *pre-shape* as it is not invariant to rotation yet. Pre-shapes lie on the complex unit sphere $\mathbb{C}S^{m-1}$. To remove rotation, all rotated versions of $\mathbf{z}$ are identified and the final shape is obtained as the resulting equivalence class of pre-shapes, denoted by $[\mathbf{z}]$. Since rotations correspond to multiplication by complex numbers of unit magnitude, the shape manifold is infact the complex projective space $\mathbb{C}P^{m-2}$. The geometry of the shape manifold is non-trivial, as a shape is represented by an equivalence class of complex vectors. This has particularly hindered the development of algorithms on the shape manifold.

It is nonetheless possible to determine the distance between two shapes. In particular, the most popular distance on the shape manifold is the full Procrustes distance. Given two pre-shapes $\mathbf{z}_1$ and $\mathbf{z}_2$, the full Procrustes distance between the corresponding shapes is given by [10, 5]

$$d_{FP}([\mathbf{z}_1],[\mathbf{z}_2]) = \left(1 - |\langle \mathbf{z}_1, \mathbf{z}_2 \rangle|^2\right)^{\frac{1}{2}}, \quad (1)$$

where $\langle .,. \rangle$ and $|.|$ denote the usual inner product in $\mathbb{C}^m$ and the absolute value of a complex number, respectively.

Under the full Procrustes distance, the Karcher mean $[\boldsymbol{\mu}]$ of a given set of pre-shapes $\mathbf{z}_1, \mathbf{z}_2, \ldots, \mathbf{z}_n$ corresponds to the dominant eigenvector of the matrix

$$\mathbf{S} = \sum_{i=1}^{n} \mathbf{z}_i \mathbf{z}_i^*. \quad (2)$$

The projection of each $\mathbf{z}_i$ to the tangent space at $[\boldsymbol{\mu}]$ is given by [5]

$$\mathbf{v}_i = \mathbf{z}_i \langle \boldsymbol{\mu}, \mathbf{z}_i \rangle - \boldsymbol{\mu} |\langle \boldsymbol{\mu}, \mathbf{z}_i \rangle|^2. \quad (3)$$

# 4. A Hilbert Space Embedding of Shapes

In this section, we discuss the advantages of embedding the shape manifold in a high-dimensional Hilbert space, and introduce a positive definite kernel on the shape manifold that makes such an embedding possible.

## 4.1. Embedding a Manifold in a Hilbert Space

Embedding lower dimensional data in a higher dimensional RKHS is commonly and successfully employed with data that lies in $\mathbb{R}^n$. The theoretical concepts of such embeddings can directly be extended to manifolds. In short, points on a manifold $\mathcal{M}$ are mapped to elements in a high (possibly infinite) dimensional Hilbert space $\mathcal{H}$, the Cauchy completion of the space spanned by real-valued functions on $\mathcal{M}$. A kernel function $k : (\mathcal{M} \times \mathcal{M}) \to \mathbb{R}$ is used to define the inner product on $\mathcal{H}$, thus making it a Reproducing Kernel Hilbert Space (RKHS). The technical difficulty in utilizing Hilbert space embeddings with manifold-valued data arises from the fact that, according to Mercer's theorem, the kernel function must be positive definite to define a valid RKHS. While many positive definite kernel functions are known for $\mathbb{R}^n$, generalizing them to manifolds is not straightforward.

Identifying such positive definite kernel functions on manifolds would, however, be greatly beneficial. Not only would it let us transform a nonlinear manifold into a linear Hilbert space, and thus allow us to exploit algorithms designed for linear spaces with manifold-valued data, but also yield a richer high dimensional representation of the original data distribution, making tasks such as classification easier.

## 4.2. A Positive Definite Kernel on the Shape Manifold

In this section, we introduce a kernel function on the shape manifold and prove its positive definiteness. Our kernel is inspired by the Gaussian radial basis function (RBF) that has proven very effective in Euclidean spaces.

More specifically, we replace the Euclidean distance in the Gaussian RBF with the full Procrustes distance on the shape manifold. This yields the kernel $k_P([\mathbf{z}_i],[\mathbf{z}_j]) := \exp(-d_{FP}^2([\mathbf{z}_i],[\mathbf{z}_j])/2\sigma^2)$, that we name *Procrustes Gaussian kernel*. While this may seem straightforward, our main contribution comes from the proof that this kernel is positive definite. Note that this is not true in general when replacing the Euclidean distance with any distance measure on shapes, as acknowledged in [26, 7, 29].

Let $\mathcal{SP}_m$ denote the $(2m-4)$ dimensional manifold of 2D shapes defined by $m$ landmarks [5]. We now present our main theorem which defines a real-valued positive definite kernel on $\mathcal{SP}_m$.

**Theorem 4.1.** *The Procrustes Gaussian kernel $k_P : (\mathcal{SP}_m \times \mathcal{SP}_m) \to \mathbb{R} : k_P([\mathbf{z}_1],[\mathbf{z}_2]) := \exp(-d_{FP}^2([\mathbf{z}_1],[\mathbf{z}_2])/2\sigma^2)$, where $d_{FP}$ is the full Procrustes distance between two shapes $[\mathbf{z}_1]$ and $[\mathbf{z}_2]$, is a positive definite kernel for all $\sigma \in \mathbb{R}$.*

*Proof.* The proof of Theorem 4.1 relies on a number of definitions, theorems and lemmas, which we state below.

We start with the definition of positive and negative definite kernels. Although common kernels used in computer vision and machine learning are real-valued, the term *positive definite kernels* covers the more general case of complex-valued kernels. We first present this more general definition [2] which will be useful for our derivation.

**Definition 4.2.** *Let $\mathcal{X}$ be a nonempty set. A kernel $f : (\mathcal{X} \times \mathcal{X}) \to \mathbb{C}$ is called **positive definite** if*

$$\sum_{i,j=1}^{n} c_i \overline{c_j} f(x_i, x_j) \geq 0$$

*for all $n \in \mathbb{N}, \{x_1, \ldots, x_n\} \subseteq \mathcal{X}$ and $\{c_1, \ldots, c_n\} \subseteq \mathbb{C}$. The kernel $f$ is called **negative definite** if it is hermitian and*

$$\sum_{i,j=1}^{n} c_i \overline{c_j} f(x_i, x_j) \leq 0$$

*for all $n \geq 2, \{x_1, \ldots, x_n\} \subseteq \mathcal{X}$ and $\{c_1, \ldots, c_n\} \subseteq \mathbb{C}$ with $\sum_{i=1}^{n} c_i = 0$.*

The following lemma lets us establish the relationship between complex and real valued positive definite kernels.

**Lemma 4.3.** *Let $\mathcal{X}$ be a nonempty set. A real-valued kernel $f : (\mathcal{X} \times \mathcal{X}) \to \mathbb{R}$ is positive (resp. negative) definite if $f$ is symmetric and $\sum_{i,j=1}^{n} c_i c_j f(x_i, x_j) \geq 0$ (resp. $\leq 0$) for all $n \in \mathbb{N}, \{x_1, \ldots, x_n\} \subseteq \mathcal{X}$ and $\{c_1, \ldots, c_n\} \subseteq \mathbb{R}$ (resp. $\sum_{i=1}^{n} c_i = 0$ in addition).*

*Proof.* See Result 3.1.6 in [2]. □

The first step of the proof of Therorem 4.1 exploits the following theorem due to Schoenberg [19].

**Theorem 4.4.** *Let $\mathcal{X}$ be a nonempty set and $f : (\mathcal{X} \times \mathcal{X}) \to \mathbb{C}$ be a kernel. The kernel $\exp(-t\, f(x_1, x_2))$ is positive definite for all $t > 0$ if and only if $f$ is negative definite.*

*Proof.* See Theorem 3.2.2 in [2]. □

This theorem lets us conclude that the positive definiteness of the Gaussian RBF kernel generated by a distance function follows from the negative definiteness of the squared distance function. Note that, in itself, this is an important result since it states the sufficient and necessary conditions to perform kernel methods on any manifold with the Gaussian kernel defined on it.

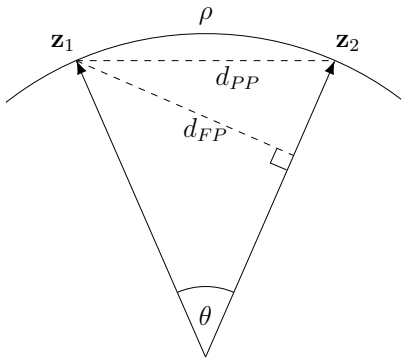

| Metric Name | Formula | Invariances | | | Pos. Def. Kernel |
|---|---|---|---|---|---|
| | | Rot. | Trans. | Scale | |
| **Full-Procrustes**, $d_{FP}$ | $\left(1 - |\langle \mathbf{z}_1, \mathbf{z}_2 \rangle|^2\right)^{\frac{1}{2}}$ | **Yes** | **Yes** | **Yes** | **Yes** |
| Partial-Procrustes, $d_{PP}$ | $\left(1 - |\langle \mathbf{z}_1, \mathbf{z}_2 \rangle|\right)^{\frac{1}{2}}$ | Yes | Yes | Partial | No |
| Arc length, $\rho$ | $\arccos(|\langle \mathbf{z}_1, \mathbf{z}_2 \rangle|)$ | Yes | Yes | Partial | No |
| Euclidean, $d_E$ | $(2 - 2\operatorname{Re}(\langle \mathbf{z}_1, \mathbf{z}_2 \rangle))^{\frac{1}{2}}$ | No | Yes | Partial | Yes |

Table 1: **Different metrics on the shape manifold**. (Left) Two pre-shapes $\mathbf{z}_1$ and $\mathbf{z}_2$ aligned in such a way that the distance between them is minimized over all possible rotations. (Right) Properties of the metrics. Note that the Euclidean distances $d_E$ is not a proper metric on the shape manifold, it is only included here for comparison purposes.

A consequence of the previous result is that we now only need to prove that the squared full Procrustes distance function on $\mathcal{SP}_m$ is negative definite.

**Lemma 4.5.** *The kernel* $f : (\mathcal{SP}_m \times \mathcal{SP}_m) \to \mathbb{R}$ : $f([\mathbf{z}_1], [\mathbf{z}_2]) := d_{FP}^2([\mathbf{z}_1], [\mathbf{z}_2]) = 1 - |\langle \mathbf{z}_1, \mathbf{z}_2 \rangle|^2$ *is negative definite.*

*Proof.* It is well-known that the kernel $g_1 : \mathbb{C}^m \times \mathbb{C}^m \to \mathbb{C}$ : $g_1(\mathbf{z}_1, \mathbf{z}_2) = \langle \mathbf{z}_1, \mathbf{z}_2 \rangle = \mathbf{z}_2^* \mathbf{z}_1$ is positive definite. It directly follows that $g_2(\mathbf{z}_1, \mathbf{z}_2) = \langle \mathbf{z}_2, \mathbf{z}_1 \rangle = \overline{\langle \mathbf{z}_1, \mathbf{z}_2 \rangle}$ is also positive definite. Since the product of two positive definite kernels is again positive definite (see Theorem 3.1.12 in [2]), $g := g_1 g_2$ is also a positive definite kernel. Now, because $g(\mathbf{z}_1, \mathbf{z}_2) = |\langle \mathbf{z}_1, \mathbf{z}_2 \rangle|^2$ is positive definite, it is easy to see that $f([\mathbf{z}_1], [\mathbf{z}_2]) = 1 - g(\mathbf{z}_1, \mathbf{z}_2)$ is negative definite from the condition that $\sum_{i=1}^n c_i = 0$ and the fact that the negative of a positive definite kernel is negative definite [2]. □

Combining Lemma 4.5 and Theorem 4.4 completes the proof of Theorem 4.1. □

Apart from the full Procrustes distance, a number of other distances have also been defined on the shape manifold. However, it can be shown using counterexamples that none of these yield positive definite Gaussian kernels of all values of $\sigma$. This result is summarized in Table 1 along with a graphical representation of different distances. In particular, we note that the rotation invariant kernel proposed in [8] uses the partial Procrustes distance and hence is not positive definite. This can be shown, for example, by considering

$\mathbf{z}_1 = [-0.387 - 0.441i, +0.501 - 0.163i, -0.114 + 0.604i]$,
$\mathbf{z}_2 = [-0.725 - 0.200i, +0.634 + 0.143i, +0.091 + 0.057i]$,
$\mathbf{z}_3 = [-0.108 - 0.373i, -0.120 - 0.411i, +0.228 + 0.784i]$,
$\mathbf{z}_4 = [+0.433 + 0.154i, -0.125 - 0.654i, -0.308 + 0.500i]$.

The rotation invariant kernel matrix computed from these pre-shapes and $\sigma = 2$ has negative eigenvalues. This proves that the rotation invariant kernel is not a valid Mercer kernel.

## 5. Kernel Methods on the Shape Manifold

The proposed Procrustes Gaussian kernel allows us to exploit algorithms designed for $\mathbb{R}^n$ on the shape manifold while accounting for the true geometry of the manifold. In this section, we briefly review the four algorithms that we use in our experiments. Note that, thanks to our kernel, any kernel-based algorithms can now be applied to data on the shape manifold. In the following, we use $k_P(.,.)$ and $\mathcal{H}_P$ to denote the Procrustes Gaussian kernel on $\mathcal{SP}_m$ and the RKHS generated by it.

### 5.1. Kernel Support Vector Machines on $\mathcal{SP}_m$

We first discuss the use of kernel SVM for binary classification on a manifold. Let $\{([\mathbf{z}_i], y_i)\}_1^n$ be the set of training examples, with $[\mathbf{z}_i] \in \mathcal{SP}_m$ and $y_i \in \{-1, 1\}$. Learning a kernel SVM consists in optimizing the parameters of a hyperplane in $\mathcal{H}_P$ so as to separate the positive and negative examples with maximum margin. At inference, this hyperplane is used to determine the class of a test point $[\mathbf{z}'] \in \mathcal{SP}_m$ mapped to $\mathcal{H}_P$, which only requires evaluating the kernel at the *support vectors* and can thus be done efficiently.

This procedure is equivalent to a standard kernel SVM formulation with a kernel matrix generated from the proposed manifold-aware kernel function. Therefore, any existing SVM software package can be employed for training and classification. Importantly, the convergence of standard SVM optimization algorithms remains guaranteed since $k_P$ is positive definite. As will be shown in our experiments, utilizing kernel SVM on the manifold yields more accurate results than existing approaches that make use of nearest-neighbor methods, kernel SVM in the tangent space, and other more sophisticated methods [8].

### 5.2. Multiple Kernel Learning on $\mathcal{SP}_m$

Multiple Kernel Learning (MKL) attempts to leverage the strengths of multiple representations of the data (e.g., image features) by combining the kernels computed from these representations. In particular, we consider the use of MKL in the context of SVM [23]. More specifically, let $\{(x_i, y_i)\}_1^n$ be the set of training examples, where $x_i$ belongs to some set $\mathcal{X}$ (e.g., $x_i$ is an image) and $y_i \in \{-1, 1\}$. Furthermore, let $\{g_j\}_1^N$, with $g_j : \mathcal{X} \to \mathcal{SP}_{m(j)}$, be a set of functions that generate different valid shapes from a given $x$. Given the kernel $k_P$, each function $g_j$ can be used to compute a kernel matrix $\mathbf{K}^{(j)}$ such that $\mathbf{K}_{pq}^{(j)} = k_P(g_j(x_p), g_j(x_q))$. These kernels are then combined to obtain a composite kernel as $\mathbf{K}^* = \sum_j \lambda_j \mathbf{K}^{(j)}$, with each $\lambda_j \geq 0$. The weights $\lambda_j$s are learnt using a min-max optimization procedure such that the resulting kernel SVM yields the best classification performance [23]. Convergence of MKL is guaranteed since all the kernels generated with $k_P$ are positive definite.

| Classification Method | Nearest Mean | Nearest Neighbor | Complex Bingham | Complex Normal | RIK | Tangent Ker. SVM | Manifold Ker. SVM |
|---|---|---|---|---|---|---|---|
| ETH-80 | 79.02 | 82.10 | 86.95 | 87.50 | 92.29 | 87.29 | **93.75** |
| COIL-100 | 72.19 | 94.76 | 91.75 | 95.47 | 95.82 | 94.44 | **97.00** |

Table 2: **Object recognition on ETH-80 and COIL-100.** Average recognition accuracies of our method compared to the state-of-the-art. Note that, on both datasets, our approach outperforms the top-performing methods presented in [8].

With data on the shape manifold, MKL provides us with a powerful tool to combine different shape representations. In particular, we generate multiple shape descriptors by varying the number of landmarks that encode a shape. The number of landmarks may affect classification accuracy: Low numbers yield representations that are robust to intra-class variations, but may not be discriminative enough to separate some classes. On the other hand, high numbers will better allow to differentiate classes, but may be more sensitive to the variations within classes.

### 5.3. Kernel PCA on $\mathcal{SP}_m$

In Euclidean spaces, kernel PCA [20] has proven effective at reducing the dimensionality of the data while accounting for its nonlinearities. With our kernel, this naturally extends to the shape manifold. More specifically, a set of shapes $\{[\mathbf{z}_i]\}_{i=1}^n$, with $[\mathbf{z}_i] \in \mathcal{SP}_m$, is mapped to the RKHS $\mathcal{H}_P$. The projection to a $d$-dimensional space is then obtained by computing the eigenvectors of the covariance matrix of the data on $\mathcal{H}_P$, which can be calculated with $k_P$. After projection, the data can be thought of as a Euclidean representation of the original manifold-valued shapes. This representation, however, was obtained by accounting for the geometry of the manifold via our kernel.

For shape retrieval from large datasets, such a Euclidean representation is highly beneficial. It allows us to perform nearest-neighbor search with the Euclidean distance in a low-dimensional Euclidean space instead of having to exploit nonlinear shape distances. As a result, shape retrieval can be made efficient using algorithms such as $k$-d trees, which cannot be utilized with nonlinear shape distances.

### 5.4. Kernel $k$-means on $\mathcal{SP}_m$

Finally, we also study the use of kernel $k$-means on the shape manifold for clustering applications. Kernel $k$-means consists in performing $k$-means clustering in a high-dimensional Hilbert space [21, 20]. More specifically, given a predefined number of clusters in $\mathcal{H}_P$, kernel $k$-means proceeds by iteratively assigning each shape $[\mathbf{z}_i] \in \mathcal{SP}_m$ of a given set $\{[\mathbf{z}_i]\}_{i=1}^n$ to its closest cluster center in $\mathcal{H}_P$, and re-computing the cluster centers as the mean of their assigned vectors in $\mathcal{H}_P$. The resulting clusters act as classes and thus allow for unsupervised shape recognition.

## 6. Experiments

We now present our experimental evaluation of the manifold-based kernel methods discussed in Section 5 on several different problems and datasets. In each experiment, we obtained the specified number of landmarks along the shape contours by uniform arc-length sampling.

### 6.1. Object Recognition

We start by tackling the problem of object recognition. To this end, we make use of ETH-80, COIL-100, MPEG-7 and Swedish leaves datasets, and employ the SVM algorithm of Section 5.1 to recognize objects using their shapes.

#### 6.1.1 ETH-80

The ETH-80 dataset [12] consists of 8 object classes (apple, car, cow, cup, dog, horse, pear, tomato), 10 different objects per class (e.g., 10 cars), and 41 images per object depicting different viewing angles and rotations. Following [8], we extracted 100 landmarks per object. We follow the standard one-vs-one kernel SVM classification procedure. The optimum value of the kernel bandwidth $\sigma$ was determined to be 0.8 by cross-validation with 1 one-vs-one classifier and was kept fixed for the other classifiers.

To compare our method with existing shape-based classification approaches, we replicated the leave-one-object-out setup of [8]: All 41 images from one object are treated as test images, and the images from the remaining 9 objects of the same class and all the other classes are used to learn classifiers. This process is repeated 80 times, once for each object, and the average accuracy is reported. In Table 2, we compare the results of our approach with the state-of-the-art results reported in [8]. Additionally, we also report results for 1NN and nearest mean classification with the Procrustes distance, as well as for tangent space SVM, for which we used the usual Euclidean Gaussian RBF kernel which is known to be positive definite. Note that our method achieves a higher accuracy than all the baselines.

| Method | Proc. 1-NN | Inner Dist. [1] | Trans. Dist. [1] | Tangent Ker. SVM | Manifold Ker. SVM |
|---|---|---|---|---|---|
| Accuracy | 91.57 | 94.70 | 95.70 | 93.57 | **96.57** |

Table 3: **Object recognition on MPEG-7.** Recognition accuracies of different methods. Note that our approach outperforms the methods in [1] and tangent space kernel SVM with the usual Gaussian RBF kernel.

### 6.1.2 COIL-100

We performed a similar experiment on the COIL-100 dataset [16] which contains 100 different objects and 72 images per object. We used 100 landmarks to represent each object, one-vs-all classifiers and the 9-fold cross-validation approach of [8]. Using validation on training data, $\sigma$ was determined to be 0.7 for all classifiers. In Table 2, we compare the classification accuracy of our method with the other state-of-the-art results reported in [8], and the baselines described in Section 6.1.1. As before, our kernel SVM on the shape manifold method attains the highest accuracy.

### 6.1.3 MPEG-7

We next evaluated our shape classifier on the MPEG-7 CE-Shape-1 dataset [11]. This dataset consists of 70 classes with 20 images per class. Since most existing algorithms for shape classification rely on nearest neighbor methods, results on this dataset are usually reported with the bull's eye score [13]. However, the bull's eye score does not apply to trained classifiers, since it assumes that all the data is available at once, instead of having train/test partitions. We therefore report results using the usual classification accuracy and compare our method with other state-of-the-art methods reporting accuracy with the same measure. Following [1], we used 10 images per class to train the classifiers and the remaining 10 for evaluation. Each shape was represented with 200 landmarks. A one-vs-all classification procedure was employed with kernel SVM. The $\sigma$ parameter, which was kept fixed for all classifiers, was set to 0.16 by cross-validation. In Table 3, we compare the results obtained with our shape manifold kernel SVM to Procrustes 1-NN classification, kernel SVM on the tangent space, and the results reported in [1]. Note that the method in [1] works in a transductive setting, and therefore has access to all the data (i.e., training and test examples) to learn a distance measure. While we make use of fewer examples to learn our classifier, our kernel allows us to achieve better accuracy.

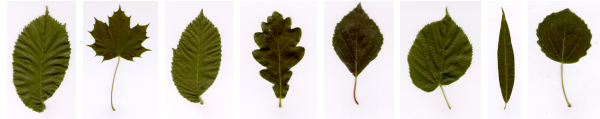

Figure 1: **Swedish leaves dataset.** Sample images from different leaf classes.

| Method | Accuracy | | |
|---|---|---|---|
| Inner Dist. [1] | 91.20 | TS MKL | 92.00 |
| Transduct. Dist. [1] | 93.80 | Manifold Ker. SVM | 91.47 |
| TS Ker. SVM | 89.47 | Manifold MKL | **94.40** |

Table 4: **Leaf identification.** Recognition accuracies on the Swedish leaves dataset.

### 6.1.4 Swedish Leaves

We now demonstrate the benefits of our kernel on the problem of leaf identification. To this end, we employed the Swedish leaves dataset [22], which contains 15 different classes with 75 images per class, 25 of which we use for training. Samples from the dataset are shown in Fig. 1. Note that some classes look very similar to each other, e.g., columns 1 & 3.

Due to the intra-class variations and inter-class similarities of this dataset, the number of landmark points used to represent the shapes affects the accuracy of the classification algorithm. We therefore used the shape manifold MKL framework proposed in Section 5.2 that allows us to better handle these variations by exploiting the shape information encoded by different number of landmarks. In particular, we computed three kernels corresponding to 200, 400 and 600 landmarks. We used one-vs-all MKL/SVM classifiers with a fixed $\sigma$ value of 0.17 for all kernels. In Table 4, we compare our MKL results with those obtained with a single kernel generated by 400 landmarks, as well as with the baselines presented in [1]. We also present the results of SVM and MKL on the tangent space (TS).

### 6.2. Shape Clustering

We next tackled the problem of visual object categorization via shape clustering. To this end, we made use of the ETH-80 dataset and employed the clustering algorithm of Section 5.4. For each object in the dataset, we used 20 images to determine the kernel bandwidth $\sigma$ and report results on the remaining images. We compare our method (Manifold KKM) against three other clustering methods on the shape manifold: $k$-means clustering on the tangent space (Tangent KM), kernel $k$-means on the tangent space with the Euclidean Gaussian kernel (Tangent KKM) and manifold $k$-means clustering (Manifold KM) that uses the Procrustes distance and Karcher mean (Section 3) at each iteration instead of the usual Euclidean distance and arithmetic mean. For each method, to overcome the sensitivity of (kernel) $k$-means to initialization, we ran the algorithm with 20 different random initializations and picked the iteration that converged to the minimum sum of point-to-centroid distances (according to the distance used by the algorithm). In Table 6, we report clustering accuracy as a function of the number of classes used. Note that our kernel $k$-means algorithm significantly outperforms the other methods.

With Matlab implementations, the times spent on clustering in the final test by Tangent KM, Tangent KKM, Manifold KM and Manifold KKM (ours) were 15.46s, 3.69s, 58.51s, and 3.61s, respectively. Note that the Manifold KM algorithm is computationally more demanding than our kernel $k$-means algorithm since it involves repeatedly computing Procrustes distances and Procrustes means.

### 6.3. Shape Retrieval

Finally, we exploited our kernel for the task of shape retrieval. State-of-the-art shape retrieval methods [13, 1] perform exhaustive nearest neighbor search over the entire database using nonlinear distances between the shapes, which does not scale with the database size. Here, instead, we make use of the kernel PCA algorithm of Section 5.3 to obtain a real valued Euclidean representation of the dataset while preserving the important shape variances. We then perform shape retrieval on the resulting Euclidean space, which can be done more efficiently. To validate our shape retrieval approach, we used the SQUID Fish dataset [14] which consists of 1100 unlabeled fish contours. We obtained 100 landmarks from each contour. We set aside 10 different fish shapes from the dataset as query images and used the remaining shapes to obtain a $d = 50$ dimensional real Euclidean representation of the dataset. For retrieval, we projected each query shape to the 50-dimensional space and found its 5 nearest neighbors using the standard Euclidean distance. Fig. 2 depicts our retrieval results. Note that the retrieved shapes match the query ones very accurately. We are unable to provide quantitative results on this dataset due to lack of ground truth.

We also evaluated our manifold kernel PCA algorithm against (non-kernel) PCA and kernel PCA on the tangent space, with the ETH dataset in the setting described in Section 6.1.1. Nearest neighbor classification on a reduced, 50-dimensional space was used with all algorithms. Table 5 summarizes the results of this experiment.

| Method | Tangent PCA | Tangent Ker. PCA | Manifold Ker. PCA |
|---|---|---|---|
| Accuracy | 75.70 | 78.14 | **81.80** |

Table 5: **Dimensionality reduction.** Recognition accuracies on the ETH-80 dataset using 1-NN in a 50-dimensional space obtained with different dimensionality reduction methods.

| Nb. of classes | Tangent KM | Tangent KKM | Manifold KM | Manifold KKM |
|---|---|---|---|---|
| 3 | 63.50 | 78.41 | 65.40 | **86.98** |
| 4 | 51.30 | 77.98 | 53.33 | **83.45** |
| 5 | 46.57 | 69.43 | 50.09 | **74.67** |
| 6 | 43.57 | 63.73 | 44.76 | **65.32** |
| 7 | 44.35 | 64.01 | 46.87 | **67.14** |
| 8 | 39.40 | 56.60 | 43.80 | **63.69** |

Table 6: **Shape clustering.** Clustering accuracies on the ETH-80 dataset. KM and KKM denote $k$-means and kernel $k$-means, respectively.

## 7. Conclusion

In this paper we have introduced a positive definite kernel on the shape manifold that allows us to embed the manifold in a high-dimensional RKHS. This Procrustes Gaussian kernel has let us extend popular kernel methods in Euclidean spaces to the shape manifold while accounting for the geometry of the manifold. To the best of our knowledge, this marks the first time popular classification methods in Euclidean space, such as SVM and MKL, have been generalized to the shape manifold. We have shown that the resulting methods achieve better performance than existing algorithms. This could be attributed to the fact that kernel methods perform recognition in a high-dimensional space while most existing methods resort to simple nearest neighbor search. In the future, we plan to extend the use of our Procrustes Gaussian kernel to other kernel algorithms.

## References

[1] X. Bai, X. Yang, L. Latecki, W. Liu, and Z. Tu. Learning Context-Sensitive Shape Similarity by Graph Transduction. *PAMI*, 2010.

[2] C. Berg, J. P. R. Christensen, and P. Ressel. *Harmonic Analysis on Semigroups*. Springer, 1984.

[3] F. L. Bookstein. Size and Shape Spaces for Landmark Data in Two Dimensions. *Statistical Science*, 1986.

[4] D. Cremers, S. J. Osher, and S. Soatto. Kernel Density Estimation and Intrinsic Alignment for Shape Priors in Level Set Segmentation. *IJCV*, 2006.

[5] I. Dryden and K. Mardia. *Statistical Shape Analysis*. Wiley, 1998.

[6] D. Gill, Y. Ritov, and G. Dror. Is Pinocchio's Nose Long or His Head Small? Learning Shape Distances for Classification. In *ISVC*, 2007.

[7] B. Haasdonk and H. Burkhardt. Invariant Kernel Functions for Pattern Analysis and Machine Learning. *Machine Learning*, 2007.

[8] O. Hamsici and A. Martinez. Rotation Invariant Kernels and Their Application to Shape Analysis. *PAMI*, 2009.

[9] S. Jayasumana, R. Hartley, M. Salzmann, H. Li, and M. Harandi. Kernel Methods on the Riemannian Manifold of Symmetric Positive Definite Matrices. In *CVPR*, 2013.

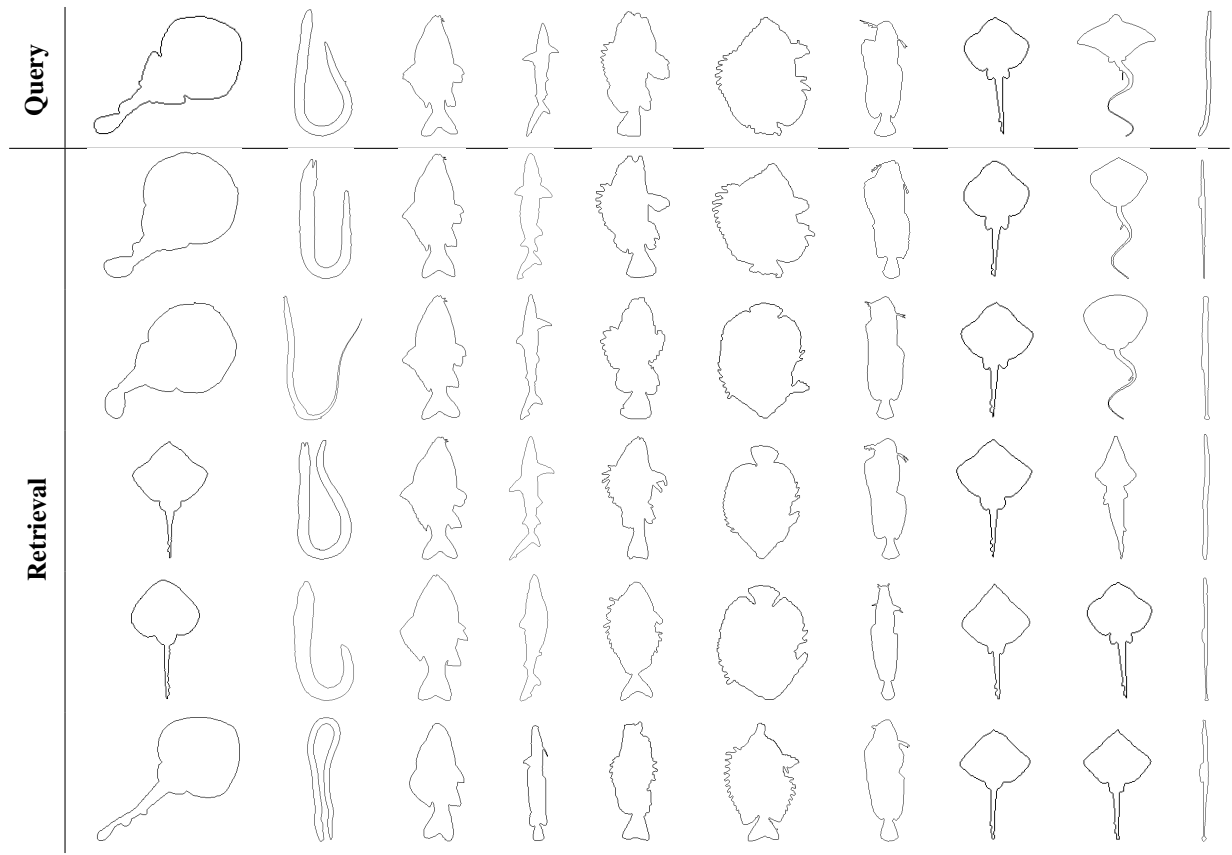

Figure 2: **Shape Retrieval.** Retrieval results from the SQUID Fish dataset using our kernel PCA approach. Nearest neighbors are ordered according to their distance to the query.

[10] D. G. Kendall. Shape Manifolds, Procrustean Metrics, and Complex Projective Spaces. *Bulletin of the London Mathematical Society*, 1984.
[11] L. Latecki, R. Lakamper, and T. Eckhardt. Shape Descriptors for Non-rigid Shapes with a Single Closed Contour. In *CVPR*, 2000.
[12] B. Leibe and B. Schiele. Analyzing Appearance and Contour Based Methods for Object Categorization. In *CVPR*, 2003.
[13] H. Ling and D. Jacobs. Shape Classification Using the Inner-Distance. *PAMI*, 2007.
[14] F. Mokhtarian, S. Abbasi, and J. Kittler. Efficient and Robust Retrieval by Shape Content through Curvature Scale Space. In *BMVC*, 1996.
[15] B. Mory, R. Ardon, A. Yezzi, and J. Thiran. Non-Euclidean Image-adaptive Radial Basis Functions for 3D Interactive Segmentation. In *ICCV*, 2009.
[16] S. A. Nene, S. K. Nayar, and H. Murase. Columbia Object Image Library (COIL-100). Technical report, 1996.
[17] C. S. Ong, S. Canu, and A. J. Smola. Learning with Non-Positive Kernels. In *ICML*, 2004.
[18] J. Platt. Fast Training of Support Vector Machines using Sequential Minimal Optimization. In *Advances in Kernel Methods — Support Vector Learning*. MIT Press, 1999.
[19] I. J. Schoenberg. Metric Spaces and Positive Definite Functions. *Transactions of the AMS*, 1938.
[20] B. Schölkopf, A. Smola, and K.-R. Müller. Nonlinear Component Analysis as a Kernel Eigenvalue Problem. *Neural Computation*, 1998.
[21] J. Shawe-Taylor and N. Cristianini. *Kernel Methods for Pattern Analysis*. Cambridge University Press, 2004.
[22] O. Söderkvist. Computer Vision Classification of Leaves from Swedish Trees. Master's thesis, Linkoping University, 2001.
[23] M. Varma and D. Ray. Learning The Discriminative Power-Invariance Trade-Off. In *ICCV*, 2007.
[24] A. Veeraraghavan, A. Roy-Chowdhury, and R. Chellappa. Matching Shape Sequences in Video with Applications in Human Movement Analysis. *PAMI*, 2005.
[25] C. Walder and O. Chapelle. Learning with Transformation Invariant Kernels. In *NIPS*, 2007.
[26] D. Weinland. Kernel Based Learning of Distances for Shape Detection. Master's thesis, University of Mannheim, 2004.
[27] G. Wu, E. Y. Chang, and Z. Zhang. An Analysis of Transformation on Non-positive Semidefinite Similarity Matrix for Kernel Machines. In *ICML*, 2005.
[28] L. Younes. Spaces and Manifolds of Shapes in Computer Vision: An Overview. *Image and Vision Comput.*, 2012.
[29] Q. Yuan, A. Thangali, V. Ablavsky, and S. Sclaroff. Learning a Family of Detectors via Multiplicative Kernels. *PAMI*, 2011.